\def\year{2022}\relax
\documentclass[letterpaper]{article} % DO NOT CHANGE THIS
\usepackage{aaai22}  % DO NOT CHANGE THIS
\usepackage{times}  % DO NOT CHANGE THIS
\usepackage{helvet}  % DO NOT CHANGE THIS
\usepackage{courier}  % DO NOT CHANGE THIS
\usepackage[hyphens]{url}  % DO NOT CHANGE THIS
\usepackage{graphicx} % DO NOT CHANGE THIS
\usepackage{natbib}  % DO NOT CHANGE THIS AND DO NOT ADD ANY OPTIONS TO IT
\usepackage{caption} % DO NOT CHANGE THIS AND DO NOT ADD ANY OPTIONS TO IT
\DeclareCaptionStyle{ruled}{labelfont=normalfont,labelsep=colon,strut=off} % DO NOT CHANGE THIS
\usepackage{algorithm}
\usepackage{algorithmic}

\usepackage{newfloat}
\usepackage{listings}
\usepackage{mathtools} % amsmath with fixes and additions
\usepackage{amsfonts}
\usepackage{amsmath}
\usepackage{amssymb}

\usepackage{yhmath}
\usepackage{enumitem}

\usepackage{hyperref}

\usepackage{booktabs} % commands to create good-looking tables
\usepackage{tikz} % nice language for creating drawings and diagrams
\newcommand{\Hquad}{\hspace{0.5em}} 
\DeclareMathOperator*{\argmin}{arg\,min}
\DeclareMathOperator*{\argmax}{arg\,max}

\usepackage{multirow}

\newcommand{\flo}[1]{\textcolor{blue}{[Flo : ]}}

\floatstyle{ruled}
\newfloat{listing}{tb}{lst}{}
\floatname{listing}{Listing}

\title{Bag Graph: Multiple Instance Learning using Bayesian Graph Neural Networks}
\author{ Soumyasundar Pal\text{*}, Antonios Valkanas\text{*}, Florence Regol, Mark Coates
}
\affiliations{
Department of Electrical and Computer Engineering, McGill University, Montreal, QC, Canada\\

\texttt{\{\href{mailto:soumyasundar.pal@mail.mcgill.ca}{soumyasundar.pal} , \href{mailto:antonios.valkanas@mail.mcgill.ca}{antonios.valkanas} , \href{mailto:florence.robert-regol@mail.mcgill.ca}{florence.robert-regol} \}@mail.mcgill.ca, \href{mailto:mark.coates@mcgill.ca}{mark.coates@mcgill.ca}}
}

\begin{document}

\maketitle
\begin{abstract}
Multiple Instance Learning (MIL) is a weakly supervised learning problem where the aim is to assign labels to sets or bags of instances, as opposed to traditional supervised learning where each instance is assumed to be {\em independent and identically distributed} (i.i.d.) and is to be labeled individually. Recent work has shown promising results for neural network models in the MIL setting. Instead of focusing on each instance, these models are trained in an end-to-end fashion to learn effective bag-level representations by suitably combining permutation invariant pooling techniques with neural architectures. In this paper, we consider modelling the interactions between bags using a graph and employ Graph Neural Networks (GNNs) to facilitate end-to-end learning. Since a meaningful graph representing dependencies between bags is rarely available, we propose to use a Bayesian GNN framework that can generate a likely graph structure for scenarios where there is uncertainty in the graph or when no graph is available. Empirical results demonstrate the efficacy of the proposed technique for several MIL benchmark tasks and a distribution regression task.
\end{abstract}
\let\thefootnote\relax\footnotetext{\\\text{*} These authors contributed equally to this work.\\
Code to reproduce our experiments is available at: \texttt{\href{https://github.com/networkslab/BagGraph}{https://github.com/networkslab/BagGraph}}}

\vspace{-1.25em}
\section{Introduction}\label{sec:intro}
\vspace{-0.25em}
% Mil vs non mil problem
In numerous supervised learning settings, we are interested in assigning a label to a {\em group} (or bag) of instances as opposed to assigning labels to the individual instances. Example application domains include drug activity prediction~\citep{musk}, disease diagnosis based on medical images~\citep{quellec2017, ilse18}, and election outcome prediction~\citep{flaxman2015}. The number of instances in each group can vary, and we often only have access to a subset of bag labels; the instances themselves do not have labels attached.

This task is known as the {\em multiple instance learning} problem. Early MIL methods such as~\citep{ramon2000} used an instance space approach where instances in each bag are processed individually and then a bag label is constructed by aggregating the instances' predictions. 
While this approach leads to explainable predictions,  it treats instances as i.i.d.\ samples from an underlying distribution.
Algorithms that make the i.i.d.\ assumption cannot model any interaction between the instances~\citep{zhou2009}, so they struggle when applied to real world problems such as medical imaging classification where strong dependencies exist and provide valuable information~\citep{quellec2017}. More recently, MIL methods have embraced bag embedding approaches~\citep{wang2018}. These methods employ some form of pooling to combine instance representations into an embedding for the entire bag.

The presence of structure between the instances in a bag motivated the use of a graph to model the dependencies. Such an approach was adopted in~\cite{zhang2011}, where a relational graph was used to specify similarities between instances. With the recent advances in graph neural networks (GNNs), there have been efforts to use these to represent the structure of instances within a bag~\citep{tu2019,yin2019}. 

Our key observation in this paper is that while graphs have been used to model the relationships between instances, they have not been employed to specify relationships between bags. In some applications, there is side-information available that provides a clear mechanism for constructing a graph. 
For example, in a real estate application when the goal is to predict mean rental prices within a neighborhood, we may assume that nearby neighborhoods tend to have similar pricing~\citep{valkanas2020}. A similar example concerns prediction of electoral results, where neighboring electoral districts are likely to exhibit similar voting patterns~\citep{flaxman2015}. A graph can then be constructed with edges representing geographic proximity. The identified dependencies are valuable in a graph-based learning framework, leading to improved predictive performance. In other cases, there is either no graph available, or the available graph information is a noisy representation of the potential relationships. Even in these circumstances it can be beneficial to explicitly learn a graph structure to represent dependencies between bags and to exploit this structure when forming label predictions.

The primary contributions of this paper are:
\begin{enumerate}%[leftmargin=*]
    % \item We formulate an end-to-end multiple instance learning architecture that incorporates (i) a set transformer to model instance interactions {\em within} bags; and (ii) a Bayesian graph neural network to jointly learn a graph topology to represent dependencies {\em between} bags and to assign labels; 
    \item We formulate an end-to-end multiple instance learning architecture that incorporates (i) existing neural network based MIL models (e.g., Deep Sets~\cite{deepsets} or Set Transformer~\cite{set_transformer}) to model instance interactions {\em within} bags; and (ii) a Bayesian graph neural network to jointly learn a graph topology to represent dependencies {\em between} bags and to assign labels; 
    
    \item We demonstrate that various instantiations of the proposed technique achieve comparable classification performance to state-of-the art methods on MIL benchmark datasets, outperform competitors in a text categorization experiment and in electoral result prediction, and offer a significant advantage in an MIL regression task.
\end{enumerate}
\vspace{-1.2em}
\section{Related Work}
\vspace{-0.25em}
The task we address can be formulated as a set learning or multiple instance learning task if we ignore the dependencies between the sets. We briefly discuss the recent related work in these fields.

{\bf Classical MIL methods} can be broadly divided into two groups: (i) instance-level and (ii) bag-level algorithms.
Instance level (or instance space) algorithms classify all individual instances within a bag, then aggregate the instance labels, and finally assign a label to the bag~\citep{ramon2000,raykar2008}. The methods can thus identify and track the instances that triggered the bag label~\citep{liu2012}. The algorithms typically rely on access to instance level training labels. Instance space methods (e.g., mi-SVM~\citep{andrews2002} and EM-DD~\citep{zhang2001}) avoid this by training models to predict instance labels without supervision. While these methods can work well, they assume that specific key instances trigger the bag label and often fail in cases where a complex relationship between instances determines the bag label.

Bag level approaches do not require access to instance labels but lack instance level explainability. 
{\em Bag space} methods (e.g.,~\citep{sun2013} and mi-Graph~\citep{zhou2009}) employ a non-vectorial distance function to compare bags. The lack of any mechanism to learn appropriate features can harm performance. {\em Embedding space} methods~(e.g.,~\cite{wang2018}, \cite{ilse18}) address this by learning fixed dimension bag embedding vectors that are used for classification. Neural networks have been applied at both the instance level~\citep{ramon2000} and more recently to derive bag embeddings~\citep{pathak2015, wang2018, ilse18}. 

{\bf Pooling}: Most embedding methods employ an encoder to embed instances to intermediate representations and then combine these through a pooling operation to obtain a bag embedding. Earlier MIL methods employed fixed (non-trainable) pooling operators, but more recently set learning and attention have been incorporated~\citep{ilse18}. \citet{set_transformer} extend this approach, using transformers and multi-head attention to learn more complex interactions between set elements.  

{\bf Graph methods for MIL}: The earliest MIL methods assumed the instances to be i.i.d., but this was relaxed in subsequent work. Indeed, it has been recognized that explicitly modeling the structure between instances and bags can be beneficial~\citep{deselaers2010}. \citet{zhang2011} employ a model where similar instances are represented as connected nodes in a relational graph. More recently, graph neural networks (GNNs) have been employed to model and learn the structure of the instances within a bag~\citep{tu2019,yin2019}.  

% Our work differs from existing work in that we represent the relationships {\em between} bags using a graph. We combine a set transformer architecture to learn the intra-bag structure  with a graph neural network to learn the inter-bag structure and train the resultant architecture in an end-to-end fashion. Furthermore, to account for scenarios where there is uncertainty in the graph or where no graph is available, we use a Bayesian graph neural network framework, jointly learning the set transformer parameters, the graph topology, and the GNN weights.  
Our work differs from existing work in that we represent the relationships {\em between} bags using a graph. We combine existing neural architectures to learn the intra-bag structure with a graph neural network to learn the inter-bag structure and train the resulting architecture in an end-to-end fashion. Furthermore, to account for scenarios where there is uncertainty in the graph or where no graph is available, we use a Bayesian graph neural network framework, jointly learning the parameters associated with the bag embedding, the graph topology, and the GNN weights.  
\vspace{-1em}
\section{Problem Statement}
\vspace{-0.25em}
We address the multiple instance learning task of mapping sets of instances (bags) to labels. Let $\mathcal{V}$ be the set of all bags. We consider a weakly supervised transductive setting, in which we observe the labels $\mathbf{y}_{\mathcal{L}} = \{\mathbf{y}_i\}_{i \in \mathcal{L}}$ for a subset of bags in a training set $\mathcal{L} \subset \mathcal{V}$. The labels $\mathbf{y}_i$ may be categorical in a classification setting or real-valued in a regression setting. 

% and are required to predict the labels of the unlabeled bags in the set $\overline{\mathcal{L}} =\mathcal{V}\setminus \mathcal{L}$

Each instance has an associated feature vector and we assume these have a common dimension, so that we can associate with each bag $i \in \mathcal{V}$ a feature matrix $\mathbf{X}_i \in \mathbb{R}^{n_i \times d_x}$, where $d_x$ is the dimensionality of each instance's feature vector and $n_i$ is the cardinality of the $i$-th bag. The number of instances can vary from bag to bag; $n_i \neq n_j \text{ for } i \neq j$. We denote the set of training features as $\mathbf{X}_{\mathcal{L}} = \{\mathbf{X}_i\}_{i \in \mathcal{L}}$. Our goal is to assign labels to the bags in the test set $\overline{\mathcal{L}} =\mathcal{V}\setminus \mathcal{L}$, for which only the features are accessible. Since we operate in a transductive setting, features from all bags $\mathbf{X}_{\mathcal{V}} = \mathbf{X}_{\mathcal{L}} \cup \mathbf{X}_{\overline{\mathcal{L}}}$ can be used during model training. 

We extend the classical MIL task by considering settings where a graph $\mathcal{G}_{obs} = (\mathcal{V},\mathcal{E})$ is provided or can be constructed through some heuristic from the available data. The nodes $i \in \mathcal{V}$ in this graph are the bags (both training and test); and an edge in the edge set $\mathcal{E}$ represents the existence of a relationship between the bags. Our method assumes that the graph is {\em homophilic}, in the sense that an edge between two nodes $i$ and $j$ is indicative of a higher probability that the bags represented by these nodes have the same label (or that the distance between the labels is small in the regression context). We consider a setting where the edges are not directed. However, the adjacency matrix can be weighted, so that the edge weights represent the varying degree of similarity between different node pairs. 

% We consider a setting where the edges are not weighted or directed, so the graph provides only qualitative information about the correlation structure rather than quantifiable metrics such as correlation coefficients.

Our problem formulation encompasses the standard MIL setting. It is equivalent to the case when the provided edge set is empty, i.e., $\mathcal{E} = \phi$. We include the subscript $obs$ in $\mathcal{G}_{obs}$ to emphasize that it is an {\em observed} graph. Our framework is designed under the assumption that there is a true, unobserved graph $\mathcal{G}$ and that the observed graph $\mathcal{G}_{obs}$ is a noisy version of this graph. We specify our adopted prior for the graph $\mathcal{G}$ and the likelihood model relating $\mathcal{G}$ and $\mathcal{G}_{obs}$ in the next section. 
\vspace{-1.25em}
\section{Methodology}
\label{sec:meth}
\vspace{-0.25em}
% Our overall learning architecture is depicted in Figure~\ref{fig:netPic}.
 We employ a Bayesian learning framework to account for uncertainties in the provided graph (or to learn it outright when one is not provided). A deep learning based MIL model is applied to the instances within each bag to generate a representation of the associated set. These representations are then aggregated using a Bayesian graph neural network to provide a final labeling for each bag. The Bayesian formulation provides a data adaptive mechanism for inferring the true graph. The architecture is trained in an end-to-end fashion, with the parameters associated with the set representation and the GNN being learned jointly with the graph topology. The loss functions are dependent on the task; we employ a cross-entropy loss for classification and mean-squared error for regression.
 
%  Within each bag a set transformer is applied to the instances to generate a representation of the associated set.

% We use SAB layers to learn an intermediate representation of the set of samples at each node. This is followed by a pooling layer using the PMA structure to obtain representations of common dimension.

% the Graph Isomorphism Network (GIN) of~\citep{gin} and

We use a typical deep-learning based MIL model which consists of two modules. First, a representation learning module is applied to the instances within each bag. This is followed by a pooling layer which summarizes the instance representations within a set to obtain a bag embedding. Subsequently, the node-level (bag-level) representations are aggregated using a GNN, which aims to take advantage of the relationships specified by the graph structure. Our framework can incorporate the vast majority of GNNs; we conduct experiments using the Graph Convolutional Network (GCN) of~\citep{kipf2017}.

Suppose that the bag representation matrix obtained from the MIL model is denoted by $\mathbf{Z}_{\mathcal{V}} \in \mathbb{R}^{|\mathcal{V}| \times d_z}$. An $L$-layer GCN uses $\mathbf{Z}_{\mathcal{V}}$ as the input and performs graph convolutions recursively as follows:
% A concrete example of a $K$-layer GCN based architecture is:
\begin{align} 
\mathbf{H}^{(1)} &= \sigma_{0}(\mathbf{\tilde{A}}\mathbf{Z}_{\mathcal{V}}\mathbf{W}^{(0)})\,,\nonumber\\
\mathbf{H}^{(\ell+1)} &= \sigma_{\ell}(\mathbf{\tilde{A}}\mathbf{H}^{(\ell)}\mathbf{W}^{(\ell)})\,, \text{ }  \ell \in \{1, 2, ... , L-1\}.\label{eq:gcn}
\end{align}
Here, $\mathbf{H}^{(\ell)} \in \mathbb{R}^{|\mathcal{V}|\times d_{\ell}}$ represents the output of $(\ell-1)$-th layer and $\mathbf{W}^{(\ell)} \in \mathbb{R}^{d_{\ell}\times d_{\ell+1}}$ is the learnable weight matrix of the $\ell$-th layer. The nonlinear activation function at the $\ell$-th layer is denoted by $\sigma_{\ell}(\cdot)$. $\mathbf{\tilde{A}} \in \mathbb{R}_{+}^{|\mathcal{V}| \times |\mathcal{V}|}$ is the non-negative, symmetric, normalized adjacency matrix of graph $\mathcal{G}$. The adjacency matrix is learned using the Bayesian framework detailed below. 

% \begin{figure*}[ht]
%     \centering
%     \includegraphics[width=15cm]{figures/networkPic2_white.png} 
%     \caption{Proposed architecture: Instance feature matrices go through the self attention block (SAB) and are pooled with multi-head attention (PMA). After the set representations are acquired we use Bayesian graph learning to model a more informative graph (with fewer inter-class edges than the graph we were given). Finally we process the network of set representations using a graph neural network.}
%     \label{fig:netPic}
% \end{figure*}
\vspace{-0.75em}
\subsection{Bayesian GNN framework}
\vspace{-0.25em}
In many graph based learning problems, the observed graph is constructed from noisy data or derived based on heuristics and/or imperfect modelling assumptions. As a result, the observed graph might not represent 
the true underlying relationship among the data on its nodes; it might contain spurious links and important links might be unobserved. However, most existing GNNs do not account for the uncertainty of the graph structure during training. 

Several recent works such as~\citep{ma2019, jiang2019, zhang2019, pal2020, elinas2020, wan2021} address this issue by incorporating probabilistic modelling or joint optimization of the graph during model training. In particular,~\citet{zhang2019} introduce a general Bayesian framework, where the observed graph is assumed to be a  random sample from a parametric random graph family and posterior inference of the true graph is considered. Despite the effectiveness of the parametric modelling approach, it has several disadvantages. The algorithm cannot be applied generally since choosing suitable random parametric models proves difficult in diverse problem settings. Posterior inference of the graph model parameters often scales poorly with the number of nodes in the graph. Finally, for many parametric random graph models (e.g. the a-MMSBM adopted in~\citep{zhang2019}), the posterior inference of the true graph cannot utilize the information provided by other known quantities such as node features and/or training labels. In order to alleviate these difficulties,~\citet{pal2020} consider a non-parametric model of the graph, which relies on a smoothness criterion of the underlying graph structure and does not impose any parametric assumptions on the graph-generative model. We adopt this approach for our GNN models. 

% Finally, if the observed graph is assumed to be a realization from a parametric random graph model, the posterior inference of the true graph cannot utilize the information provided by other known quantities such as node features and/or training labels \MJC{Maybe rewrite this sentence - an exponential random graph model could be dependent on these aspects - it's only because Zhang et al.\ adopts a model (SBM) where the likelihood does not depend on these}.

In the Bayesian setting, the task is to approximate the posterior distribution of the unknown test set labels $\mathbf{y}_{\overline{\mathcal{L}}}$ conditioned on the training labels $\mathbf{y}_{\mathcal{L}}$, the node (bag) features $\mathbf{X}_{\mathcal{V}} = \{\mathbf{X}_i\}_{i \in {\mathcal{V}}}$, and (possibly) the observed graph $\mathcal{G}_{obs}$. This can be represented by computing the expectation of the model likelihood w.r.t. the posterior distributions of the true graph $\mathcal{G}$, the GNN weights $\mathbf{W} = \{\mathbf{W}^{(\ell)}\}_{\ell=0}^{L-1}$ and the MIL model parameters $\Theta$ as follows:
\begin{align}
&p(\mathbf{y}_{\overline{\mathcal{L}}}|\mathbf{y}_{\mathcal{L}},\mathbf{X}_{\mathcal{V}},\mathcal{G}_{obs}) = \int p(\mathbf{y}_{\overline{\mathcal{L}}}|\mathbf{W},\mathcal{G},\mathbf{Z}_{\mathcal{V}})p(\mathbf{W}|\mathbf{y}_{\mathcal{L}},\mathbf{Z}_{\mathcal{V}},\mathcal{G})\,\nonumber\\
&p(\mathcal{G}|\mathcal{G}_{obs}, \mathbf{Z}_{\mathcal{V}}, \mathbf{y}_{\mathcal{L}}) p(\mathbf{Z}_{\mathcal{V}}|\mathbf{X}_{\mathcal{V}}, \Theta) p(\Theta) \,d\Theta \,d\mathbf{Z}_{\mathcal{V}} \,d\mathbf{W}\,d\mathcal{G} \,. \label{eq:exact_posterior}
\end{align}
Here, $p(\Theta)$ is the prior distribution of the MIL model parameters and $p(\mathbf{Z}_{\mathcal{V}}|\mathbf{X}_{\mathcal{V}}, \Theta)$ represents the deterministic operation to obtain the bag representation matrix $\mathbf{Z}_{\mathcal{V}}$, which is used as an input to the Bayesian GNN. We approximate the integral over $\Theta$ and $\mathbf{Z}_{\mathcal{V}}$ by maximum likelihood estimates:
\begin{align}
\widehat{\mathbf{Z}}_{\mathcal{V}} = \text{MIL}\big(\mathbf{X}_{\mathcal{V}}, \widehat{\Theta}\big)\,, \label{eq:ML_estimate}
\end{align}
where $\widehat{\Theta}$ and $\widehat{\mathbf{Z}}_{\mathcal{V}}$ are the ML estimates. In a classification problem, the likelihood $p(\mathbf{y}_{\overline{\mathcal{L}}}|\mathbf{W},\mathcal{G},\widehat{\mathbf{Z}}_{\mathcal{V}})$ of the test set labels is a categorical distribution which can be modelled by applying a softmax function to the output $\mathbf{H}^{(L)}$ of the last layer of the GNN. A Gaussian likelihood can be used in the regression setting. Since the integral in eq.~\eqref{eq:exact_posterior} is intractable, a Monte Carlo approximation is formed as follows:
\begin{align}
p(\mathbf{y}_{\overline{\mathcal{L}}}|\mathbf{y}_{\mathcal{L}},\mathbf{X}_{\mathcal{V}},\mathcal{G}_{obs})  \approx 
\dfrac{1}{S} \sum_{s=1}^S
p(\mathbf{y}_{\overline{\mathcal{L}}}|\mathbf{W}_{s},\widehat{\mathcal{G}},\widehat{\mathbf{Z}}_{\mathcal{V}})\,.
\label{eq:MC_posterior}
\end{align}
Here, $\widehat{\mathcal{G}} = \displaystyle{\argmax_{\mathcal{G}}}\Hquad p(\mathcal{G}|\mathcal{G}_{obs}, \widehat{\mathbf{Z}}_{\mathcal{V}}, \mathbf{y}_{\mathcal{L}})$ denotes the {\em maximum a posteriori} (MAP) estimate of the true graph $\mathcal{G}$. 
The posterior of the GNN weights $p(\mathbf{W}|\mathbf{y}_{\mathcal{L}},\widehat{\mathbf{Z}}_{\mathcal{V}},\widehat{\mathcal{G}})$ is approximated by training a Bayesian GNN using the graph $\widehat{\mathcal{G}}$ and sampling $S$ weight matrices $\{\mathbf{W}_{s}\}_{s=1}^S$ using MC dropout~\citep{gal2016}. This is equivalent to sampling $\mathbf{W}_{s}$ from a particular variational approximation of the true posterior of the weights, if the prior distribution $p(\mathbf{W})$ is Gaussian.

In the non-parametric graph generative model described in~\citet{pal2020}, the undirected random graph $\mathcal{G}$ is specified in terms of its symmetric adjacency matrix $\mathbf{A}_{\mathcal{G}} \in \mathbb{R}_+^{|\mathcal{V}| \times |\mathcal{V}|}$. The prior distribution for $\mathcal{G}$ ensures that there is no disconnected node in $\mathcal{G}$ and it is not extremely sparse. 
% \MJC{Is it really jsut ensuring that the edge weights are small? Can this be explained more clearly? - It's not obvious why small edge weights would be a desirable thing.}
\begin{align}
  p(\mathcal{G}) \propto
  \begin{dcases}
    \exp{\big(\alpha\mathbf{1}^{\top}\log(\mathbf{A}_{\mathcal{G}}\mathbf{1}) - \beta \|\mathbf{A}_{\mathcal{G}} \|_F^2\big)}\,, &\text{if }  \mathbf{A}_{\mathcal{G}} \geqslant \mathbf{0}, \\
      &\text{\phantom{if}} \mathbf{A}_{\mathcal{G}} = \mathbf{A}_{\mathcal{G}}^{\top}\\
    0\,, &  \text{otherwise}\,.
    \end{dcases}
\end{align}
Here, $\|\cdot\|_F$ denotes the Frobenius norm and the hyperparameters $\alpha$ and $\beta$ control the scale and sparsity of $\mathbf{A}_{\mathcal{G}}$.
The joint likelihood of $\mathcal{G}_{obs}$, $\widehat{\mathbf{Z}}_{\mathcal{V}}$, and $\mathbf{y}_{\mathcal{L}}$ encourages higher edge weights for similar node pairs and lower edge weights for dissimilar node pairs. The functional form of the likelihood is specified as:
\vspace{-0.25em}
 \begin{align}
    p(\mathcal{G}_{obs}, \widehat{\mathbf{Z}}_{\mathcal{V}}, \mathbf{y}_{\mathcal{L}}|\mathcal{G}) \propto \exp{\big(- \|\mathbf{A}_{\mathcal{G}} \circ \mathbf{D}(\mathcal{G}_{obs}, \widehat{\mathbf{Z}}_{\mathcal{V}}, \mathbf{y}_{\mathcal{L}}) \|_{1,1}\big)}\,.\label{eq:joint_likelihood}
\end{align}
Here, 
$\circ$ indicates the Hadamard product and $\|\cdot\|_{1,1}$ stands for the elementwise $\ell_1$ norm. $\mathbf{D}(\mathcal{G}_{obs}, \widehat{\mathbf{Z}}_{\mathcal{V}}, \mathbf{y}_{\mathcal{L}}) \geqslant \mathbf{0}$ is a non-negative, symmetric pairwise distance matrix which measures the dissimilarity between the nodes. 
We have:
\begin{align}
 \mathbf{D}_{ij} (\mathcal{G}_{obs}, \widehat{\mathbf{Z}}_{\mathcal{V}}, \mathbf{y}_{\mathcal{L}}) &= \text{dist}(\boldsymbol{z}_i, \boldsymbol{z}_j)\,,\label{eq:dist}
\end{align}
where, $\boldsymbol{z}_i$ denotes some representation of node $i$ and $\text{dist}(\cdot, \cdot)$ is a distance metric. In our experiments, we form $\mathbf{D}$ by computing the pairwise squared Euclidean distance between the bag representations from the last layer of a base model $\widehat{\mathbf{y}}_{\overline{\mathcal{L}}} = f_{\phi} (\mathbf{X}_{\mathcal{V}}, \mathbf{y}_{\mathcal{L}}, \mathcal{G}_{obs})$, (e.g. an end-to-end deep-learning based MIL model or an MIL model combined with a GNN trained on the observed graph $\mathcal{G}_{obs}$). This flexibility in construction of the distance matrix $\mathbf{D}$ allows the application of our Bayesian approach to settings where $\mathcal{G}_{obs}$ is not available. It also proves useful in cases where we only have access to a heuristically constructed $\mathcal{G}_{obs}$, which poorly expresses the true relationships between bags. 

Instead of sampling $\mathcal{G}$ from a high dimensional posterior distribution ($\mathcal{O}(|\mathcal{V}|^2)$, where $|\mathcal{V}|$ is the number of the nodes), we adopt a MAP estimation approach as in~\cite{pal2020}. We estimate the graph as:
\vspace{-0.25em}
 \begin{align}
 \widehat{\mathcal{G}} &= \argmax_{\mathcal{G}} p(\mathcal{G}|\mathcal{G}_{obs}, \widehat{\mathbf{Z}}_{\mathcal{V}}, \mathbf{y}_{\mathcal{L}})\,,\label{opt:graph_inference}
 \end{align}
 \vspace{-0.25em}
Solving this is equivalent to learning a $|\mathcal{V}|\!\times\!|\mathcal{V}|$ non-negative, symmetric adjacency matrix of $\widehat{\mathcal{G}}$. We can re-express the optimization task as:
\begin{align}
      \label{eq:adj_opt}
  \mathbf{A}_{\widehat{\mathcal{G}}}&= \argmin_{\substack{\mathbf{A}_{\mathcal{G}} \in \mathbb{R}_+^{|\mathcal{V}|\times |\mathcal{V}|}, \\ \mathbf{A}_{\mathcal{G}}=\mathbf{A}_{\mathcal{G}}^{\top} }}
  \begin{aligned}[t]
  &\|\mathbf{A}_{\mathcal{G}} \circ \mathbf{D}\|_{1,1} -\alpha\mathbf{1}^{\top}\log(\mathbf{A}_{\mathcal{G}}\mathbf{1}) \\
  & \quad \quad \quad \quad \quad \quad \quad + \beta\|\mathbf{A}_{\mathcal{G}}\|_F^2 \,.
 \end{aligned}
 \end{align}

\citet{kalofolias2016} uses a primal-dual optimization algorithm to solve this problem in the context of learning a graph from smooth signals. In this work, we use the approximate algorithm in~\cite{kalofolias2019}. This algorithm allows for a favourable computational complexity for the graph inference and provides a useful heuristic for hyperparameter selection. The overall algorithm is summarized in Algorithm~\ref{alg:bgcn_mil}.
% \MJC{The ``suitable base model'', ``suitable prior distribution'', and ``suitable loss function'' are wishy-washy. Maybe we can just specify that these are inputs - base model $f_\phi$, prior $p(W)$, and loss $\mathcal{L}$? Then we could include a paragraph to discuss these inputs and perhaps provide some examples?}
\begin{algorithm}[ht]
\footnotesize
\caption{MIL using Bayesian GNN with non-parametric graph learning}
\label{alg:bgcn_mil}
% \scriptsize
\begin{algorithmic}[1]
\STATE {\bfseries Input:}   $\mathbf{X}_{\mathcal{V}}$, $\mathbf{y}_{\mathcal{L}}$, and $\mathcal{G}_{obs}$
\STATE {\bfseries Output:}  $p(\mathbf{y}_{\overline{\mathcal{L}}}|\mathbf{y}_{\mathcal{L}},\mathbf{X}_{\mathcal{V}},\mathcal{G}_{obs})$

\STATE Train a base model $f_{\phi}$ using $\mathbf{X}_{\mathcal{V}}$, $\mathbf{y}_{\mathcal{L}}$, and (possibly) $\mathcal{G}_{obs}$ to learn $\boldsymbol{z}_i$ for $1 \leqslant i \leqslant |\mathcal{V}|$. Compute $\mathbf{D}$ using eq.~\eqref{eq:dist}. 

\STATE  Solve the optimization problem in~\eqref{eq:adj_opt} to obtain $\mathbf{A}_{\widehat{\mathcal{G}}}$ (equivalently, $\widehat{\mathcal{G}}$).

\STATE Assuming a Gaussian prior distribution $p(\mathbf{W})$ for $\mathbf{W}$, train the MIL model combined with GNN over the graph $\widehat{\mathcal{G}}$ using a suitable loss function $\mathcal{L}(\widehat{\mathbf{y}}_{\mathcal{L}}, \mathbf{y}_{\mathcal{L}})$ to optimize $\Theta$ and $\mathbf{W}$ jointly.

\STATE Keeping $\Theta$ fixed at the learned value $\widehat{\Theta}$, obtain $\widehat{\mathbf{Z}}_{\mathcal{V}} = \text{MIL} \big(\mathbf{X}_{\mathcal{V}}, \widehat{\Theta}\big)$.

\FOR{$s=1$ {\bfseries to} $S$}
\STATE Apply MC dropout in the GNN layers to sample $\mathbf{W}_{s}$.
\ENDFOR

\STATE Approximate $p(\mathbf{y}_{\overline{\mathcal{L}}}|\mathbf{y}_{\mathcal{L}},\mathbf{X}_{\mathcal{V}},\mathcal{G}_{obs})$ using~\eqref{eq:MC_posterior}.
\end{algorithmic}
\end{algorithm}
\begin{table}[ht]
\centering
\caption{Mean and standard error (when available) of classification accuracy (in \%) for benchmark MIL datasets. The best and the second best results in each column are shown in bold and marked with underline respectively. Higher accuracies are better.}
\label{tab:mil_data}
\vspace{-0.5em}
\scriptsize
\setlength{\tabcolsep}{2pt}
\begin{tabular}{lccccc}
\toprule % from booktabs package
\textbf{Algorithm} &\textbf{MUSK1} &\textbf{MUSK2} &\textbf{FOX} &\textbf{TIGER} &\textbf{ELEPHANT} \\
\midrule % from booktabs package
mi-SVM &87.4$\pm$N/A &83.6$\pm$N/A &58.2$\pm$N/A  &78.4$\pm$N/A  &82.2$\pm$N/A  \\
MI-SVM &77.9$\pm$N/A &84.3$\pm$N/A &57.8$\pm$N/A &84.2$\pm$N/A &84.3$\pm$N/A   \\
MI-Kernel &88.0$\pm$3.1 &\underline{89.3$\pm$1.5} &60.3$\pm$2.8 &84.2$\pm$1.0 &84.3$\pm$1.6           \\
EM-DD &84.9$\pm$4.4 &86.9$\pm$4.8 &60.9$\pm$4.5 &73.0$\pm$4.3 &77.1$\pm$4.3  \\
mi-Graph &88.9$\pm$3.3 &\textbf{90.3$\pm$3.9} &62.0$\pm$4.4 &\textbf{86.0$\pm$3.7} &86.9$\pm$3.5  \\
MI-VLAD &87.1$\pm$4.3 &87.2$\pm$4.2 &62.0$\pm$4.4 &81.1$\pm$3.9 &85.0$\pm$3.6  \\
mi-FV &\textbf{90.9$\pm$4.2} &88.4$\pm$4.2 &62.1$\pm$4.9 &81.3$\pm$3.7 &85.2$\pm$3.6  \\
\midrule
mi-Net &88.9$\pm$3.9 &85.8$\pm$4.9 &61.3$\pm$3.5 &82.4$\pm$3.4 &85.8$\pm$3.7 \\
MI-Net &88.7$\pm$4.1  &85.9$\pm$4.6 &62.2$\pm$3.8 &83.0$\pm$3.2 &86.2$\pm$3.4 \\
MI-Net (DS) &89.4$\pm$4.2  &87.4$\pm$4.3 &\underline{63.0$\pm$3.7} &\underline{84.5$\pm$3.9} &\underline{87.2$\pm$3.2}   \\
MI-Net (RC) &89.8$\pm$4.3 &87.3$\pm$4.4 &61.9$\pm$4.7 &83.6$\pm$3.7 &85.7$\pm$4.0  \\
Attention &89.2$\pm$4.0  &85.8$\pm$4.8 &61.5$\pm$4.3 &83.9$\pm$2.2 &86.8$\pm$2.2   \\
Gated-Attention &\underline{90.0$\pm$5.0} &86.3$\pm$4.2 &60.3$\pm$2.9 &\underline{84.5$\pm$1.8} &85.7$\pm$2.7 \\
\midrule
rFF+pool     &88.7$\pm$3.7 &87.1$\pm$3.8  &61.1$\pm$4.1 &82.8$\pm$2.1  &\textbf{87.5$\pm$3.0}  \\
rFF+pool-GCN &89.9$\pm$3.0  &86.0$\pm$4.1  &62.9$\pm$3.4 &82.9$\pm$2.2  &\textbf{87.5$\pm$3.0}   \\
B-rFF+pool-GCN  &89.9$\pm$3.6 &87.2$\pm$2.6 &\textbf{63.9$\pm$2.7} &83.0$\pm$2.1  &84.2$\pm$3.4  \\ 
\bottomrule
\end{tabular}
\vspace{-2.5em}
\end{table}
\begin{table*}[ht]
\footnotesize
\centering
\caption{Mean and std. error (when available) of classification accuracy (in \%) along with average and median ranks (lower ranks are better) of the algorithms for the 20 text categorization datasets derived from the 20 Newsgroups corpus. %The best and the second best results in each row are shown in bold and marked with underline respectively. 
Higher accuracies and lower ranks are better.}
\label{tab:text}
\vspace{-0.5em}
\scriptsize
\setlength{\tabcolsep}{4pt}
\begin{tabular}{l|ccc||cccc||ccc}
\hline 
\multirow{2}{*}{\textbf{Algorithm}} &\multirow{2}{*}{\begin{tabular}[c]{@{}c@{}}\textbf{MI-}\\ \textbf{Kernel}\end{tabular}} &\multirow{2}{*}{\begin{tabular}[c]{@{}c@{}}\textbf{mi-}\\ \textbf{Graph}\end{tabular}} &\multirow{2}{*}{\begin{tabular}[c]{@{}c@{}}\textbf{mi-}\\ \textbf{FV}\end{tabular}} &\multirow{2}{*}{\begin{tabular}[c]{@{}c@{}}\textbf{mi-}\\ \textbf{Net}\end{tabular}} &\multirow{2}{*}{\begin{tabular}[c]{@{}c@{}}\textbf{MI-}\\ \textbf{Net}\end{tabular}} &\multirow{2}{*}{\begin{tabular}[c]{@{}c@{}}\textbf{MI-}\\ \textbf{Net (DS)}\end{tabular}} &\multirow{2}{*}{\begin{tabular}[c]{@{}c@{}}\textbf{MI-}\\ \textbf{Net (RC)}\end{tabular}} &\multirow{2}{*}{\textbf{Res+pool}} &\multirow{2}{*}{\begin{tabular}[c]{@{}c@{}}\textbf{Res+pool-}\\ \textbf{GCN} \end{tabular}} &\multirow{2}{*}{\begin{tabular}[c]{@{}c@{}}\textbf{B-Res+pool-}\\ \textbf{GCN}\end{tabular}} \\ 
&  &  &  &  &  &  &  &  &  & \\ \hline 

Average rank  &10.00    &8.70   &7.50   &4.60   &\underline{3.70}   &4.05  &4.50   &4.05  &4.55  &\textbf{3.35} \\ 
Median rank  &10.00   &9.00   &8.00   &5.00   &4.00   &4.00   &4.00   &\underline{3.50}  &4.50  &\textbf{2.50} \\ \hline \hline                                           
\emph{alt.atheism}  &60.2$\pm$3.9 &65.5$\pm$4.0 &84.8 &83.1$\pm$2.3 &84.7$\pm$1.8 &84.4$\pm$2.0 &83.6$\pm$1.5  &\underline{88.3$\pm$2.2}   &87.6$\pm$2.7   &\textbf{88.8$\pm$2.0} \\

\emph{comp.graphics}  &47.0$\pm$3.3 &77.8$\pm$1.6 &59.4 &81.7$\pm$0.6 &\textbf{82.0$\pm$1.5} &\underline{81.9$\pm$0.5} &81.5$\pm$0.9  &80.0$\pm$3.2   &78.7$\pm$2.3   &79.8$\pm$3.2 \\

\emph{comp.os.ms-windows.misc}  &51.0$\pm$5.2 &63.1$\pm$1.5 &61.5 &70.4$\pm$1.7 &70.7$\pm$1.1 &70.9$\pm$1.1 &70.7$\pm$1.4  &\textbf{71.7$\pm$3.6}   &\underline{71.1$\pm$3.9}   &70.3$\pm$3.8  \\ 

\emph{comp.sys.ibm.pc.hardware} &46.9$\pm$3.6 &59.5$\pm$2.7 &66.5 &\textbf{79.0$\pm$1.8} &\underline{78.6$\pm$1.0} &78.3$\pm$1.3 &78.5$\pm$1.0  &73.1$\pm$3.4   &73.0$\pm$2.9   &75.8$\pm$3.8 \\

\emph{comp.sys.mac.hardware} &44.5$\pm$3.2 &61.7$\pm$4.8  &66.0 &\underline{79.4$\pm$1.6} &79.1$\pm$1.5 &\textbf{79.7$\pm$1.1} &79.2$\pm$1.9  &79.3$\pm$3.1   &78.2$\pm$2.6   &78.7$\pm$3.3 \\

\emph{comp.windows.x} &50.8$\pm$4.3 &69.8$\pm$2.1 
&76.8 &79.9$\pm$1.8 &80.9$\pm$1.9 &80.1$\pm$1.1 &81.2$\pm$2.7 &84.9$\pm$2.7   &\underline{85.7$\pm$2.9}   &\textbf{86.1$\pm$1.9}  \\

\emph{misc.forsale} &51.8$\pm$2.5 &55.2$\pm$2.7  &56.5 &67.1$\pm$0.9 &66.7$\pm$1.2 &66.0$\pm$1.6 &67.2$\pm$1.2 &\textbf{75.8$\pm$3.5}   &74.0$\pm$3.6   &\underline{74.4$\pm$3.6}  \\

\emph{rec.autos} &52.9$\pm$3.3 &72.0$\pm$3.7
&66.7 &76.5$\pm$1.2 &76.9$\pm$1.6 &76.4$\pm$1.6 &76.1$\pm$1.6 &78.3$\pm$3.3   &\textbf{78.8$\pm$2.8}   &\underline{78.5$\pm$3.2} \\

\emph{rec.motorcycles} &50.6$\pm$3.5 &64.0$\pm$2.8 &80.2 &83.4$\pm$1.1 &84.2$\pm$1.0 &83.5$\pm$1.5 &83.3$\pm$1.3 &\underline{85.0$\pm$2.4}   &84.8$\pm$2.9   &\textbf{85.8$\pm$2.5}   \\

\emph{rec.sport.baseball} &51.7$\pm$2.8 &64.7$\pm$3.1 &77.9 &86.0$\pm$1.6 &\underline{86.7$\pm$1.7} &85.7$\pm$2.5  &\textbf{87.1$\pm$1.4} &80.0$\pm$3.1   &81.4$\pm$3.6   &83.4$\pm$4.1 \\

\emph{rec.sport.hockey} &51.3$\pm$3.4 &85.0$\pm$2.5 &82.3 &89.0$\pm$1.7 &\underline{90.2$\pm$1.4} &\textbf{91.1$\pm$1.6} &89.8$\pm$1.1 &89.9$\pm$2.3   &89.4$\pm$2.9   &90.0$\pm$2.9 \\

\emph{sci.crypt} &56.3$\pm$3.6 &69.6$\pm$2.1
&76.0 &79.5$\pm$1.4 &77.9$\pm$1.5 &77.8$\pm$2.6 &78.6$\pm$2.3  &80.1$\pm$3.7   &\underline{81.8$\pm$3.0}   &\textbf{81.9$\pm$3.4} \\

\emph{sci.electronics} &50.6$\pm$2.0 &87.1$\pm$1.7  &55.5 &92.1$\pm$0.8 &\textbf{93.2$\pm$0.4} &92.7$\pm$0.5 &\underline{93.1$\pm$0.7}  &90.4$\pm$2.9   &90.7$\pm$3.0   &91.4$\pm$2.8  \\

\emph{sci.med} &50.6$\pm$1.9 &62.1$\pm$3.9 
&78.3 &\textbf{85.5$\pm$0.9} &84.2$\pm$0.7 &\underline{84.7$\pm$1.3} &83.8$\pm$1.4 &78.4$\pm$3.2   &78.5$\pm$2.5   &80.2$\pm$3.0 \\

\emph{sci.space} &54.7$\pm$2.5 &75.7$\pm$3.4 
&81.8 &79.8$\pm$1.3 &79.5$\pm$2.8 &80.1$\pm$2.6 &80.3$\pm$2.6 &88.1$\pm$2.6   &\underline{88.3$\pm$2.8}   &\textbf{88.9$\pm$2.9} \\

\emph{soc.religion.christian} &49.2$\pm$3.4 &59.0$\pm$4.7  &\textbf{81.4} &79.9$\pm$1.5 &\underline{80.7$\pm$1.7} &80.1$\pm$1.4 &80.5$\pm$2.0 &78.7$\pm$3.6   &78.1$\pm$3.3   &79.4$\pm$2.0 \\

\emph{talk.politics.guns} &47.7$\pm$3.8 &58.5$\pm$6.0  &74.7 &76.1$\pm$1.9 &\textbf{78.2$\pm$1.8} &77.0$\pm$2.4 &77.3$\pm$1.0  &76.0$\pm$4.9   &73.6$\pm$3.7   &\underline{77.7$\pm$4.2}   \\

\emph{talk.politics.mideast} &55.9$\pm$2.8 &73.6$\pm$2.6 &79.3 &\underline{83.9$\pm$1.0} &\textbf{84.0$\pm$1.2} &83.8$\pm$1.0 &83.3$\pm$2.0 &82.1$\pm$3.4   &81.4$\pm$3.9   &81.6$\pm$3.1 \\

\emph{talk.politics.misc} &51.5$\pm$3.7 &70.4$\pm$3.6  &69.7 &76.5$\pm$1.5 &75.8$\pm$2.3 &\underline{76.8$\pm$2.2} &75.6$\pm$1.9 &76.5$\pm$5.0   &\underline{76.8$\pm$4.6}   &\textbf{77.9$\pm$4.7}  \\

\emph{talk.religion.misc} &55.4$\pm$4.3 &63.3$\pm$3.5 &73.9 &74.4$\pm$1.5 &76.2$\pm$1.7 &76.2$\pm$1.5 &74.3$\pm$1.2 &\underline{79.0$\pm$3.6}   &78.8$\pm$4.3   &\textbf{80.0$\pm$3.7}  \\ \hline
\end{tabular}
\vspace{-2.5em}
\end{table*}

\vspace{-1.5em}
\section{Experiments}
\label{sec:experiment}
\vspace{-0.25em}
We perform classification experiments on 5 benchmark MIL datasets, 20 text datasets from the 20Newsgroups corpus, and the 2016 US election data. In addition, we also consider a distribution regression task of predicting neighborhood property rental prices in New York City.
\vspace{-0.5em}
\subsection{Classification of benchmark MIL datasets}
\label{subsec:benchmark}
\vspace{-0.25em}
In this experiment, we evaluate the proposed approach on five popular benchmark datasets: MUSK1, MUSK2~\cite{musk}, FOX, TIGER, and ELEPHANT~\cite{andrews2002}. The detailed description of these datasets are provided in the supplementary.  

% The MUSK1 and MUSK2~\cite{musk} datasets are composed of 92 and 102 molecules, respectively. Some of the molecules have a musky smell while others do not. The goal is to predict whether the molecules will be musky or non-musky. A bag is constructed for each molecule with its various geometric arrangements, called conformations, as instances. Each instance is represented with a 166 dimensional feature vector. If a molecule has at least one musky conformation, it is labeled as musky. The FOX, TIGER, and ELEPHANT~\cite{andrews2002} datasets consist of 200 bags of features extracted from animal images. A bag is labeled positive if at least one instance contains the relevant animal whereas the negative bags contain images of different animals. In these datasets, each image is associated with a 230 dimensional feature vector. 

We compare our approach with the following baselines:\\
\textbf{Instance space methods:} mi-SVM and MI-SVM~\citep{andrews2002}, EM-DD~\citep{zhang2001}, MI-VLAD and mi-FV~\citep{wei2017}.\\
\textbf{Bag space methods:} MI-Kernel~\citep{gartner2002}, mi-Graph~\citep{zhou2009}.\\
\textbf{Embedding space methods:} mi-Net and MI-Net~\citep{wang2018}, Attention Neural Network and Gated Attention Neural Network~\citep{ilse18}: These methods use neural networks and attention to learn embeddings of the bags.
% \textbf{Instance-level GNN:}~\cite{tu2019} use a GNN to learn the structural relationships between instances within a bag.

In order to instantiate the proposed approach, we adopt the following procedure. We first design a suitable graph agnostic, deep learning based MIL algorithm as a base model. For this experiment, we consider a row-wise FeedForward architecture with pooling (rFF+pool)~\cite{deepsets} as the base model. We equip this architecture with deep supervision~\cite{wang2018}. Next, we tune the model based on a 10 fold cross-validation. Once the architecture and other hyperparameters such as learning rate, number of training epochs, and weight decay are fixed, we only replace the last linear layer of the base model with a GCN layer to form a GCN variant. The proposed Bayesian approach uses the same architecture. This approach ensures that the GCN and the BGCN variants have the same number of learnable parameters, the same hyperparameters, and similar training complexity as the base model. Moreover, the only difference between the GCN and the BGCN variants is that the GCN uses an observed graph $\mathcal{G}_{obs}$, whereas the Bayesian approach estimates $\widehat{\mathcal{G}}$ form the data. 

Since no graph is specified for these datasets, we apply a heuristic to create the observed graph $\mathcal{G}_{obs}$. We follow a simple $k$-nearest-neighbor approach, evaluating the distance between bags as the Euclidean distance between the embeddings obtained from the base model. Edges are added between nodes with nearby embeddings, with each node adding an edge to its nearest $k$ neighbors. For the proposed Bayesian approach, we have two hyperparameters $k$ and $r$ associated with the approximate graph inference technique in~\cite{kalofolias2019}, used in Step 4 of Algorithm~\ref{alg:bgcn_mil}. A permissible edge set is first constructed based on a $kr$-NN graph which greatly alleviates the computational complexity of the graph learning algorithm. Subsequently, a primal-dual algorithm is run on this reduced edge set to obtain $\widehat{\mathcal{G}}$, in which each node has approximately $k$ neighbors. We choose these hyper-parameters using 10 fold cross-validation. The detailed description of the architecture and the hyperparameters are summarized in the supplementary. These general steps are also followed in the other experiments.

We perform 10-fold cross validation for 10 times with different random data partitions and report the mean accuracy with its standard error in Table~\ref{tab:mil_data}. \citet{ilse18} remark that deep learning approaches are not well suited for these datasets as they are composed of precomputed features and the cardinalities of the bags are relatively small. From Table~\ref{tab:mil_data}, we observe that the base model rFF+pool achieves comparable performance to the neural network based approaches (note the standard errors of the mean accuracies). The rFF+pool-GCN and the proposed B- rFF+pool-GCN offer a relatively small improvement in accuracy compared to the base model in most cases.
\begin{figure}[ht]
% \centering
\hspace{-1em}
\includegraphics[trim=6cm 0.2cm 0cm 0.25cm, clip, width=0.5\textwidth]{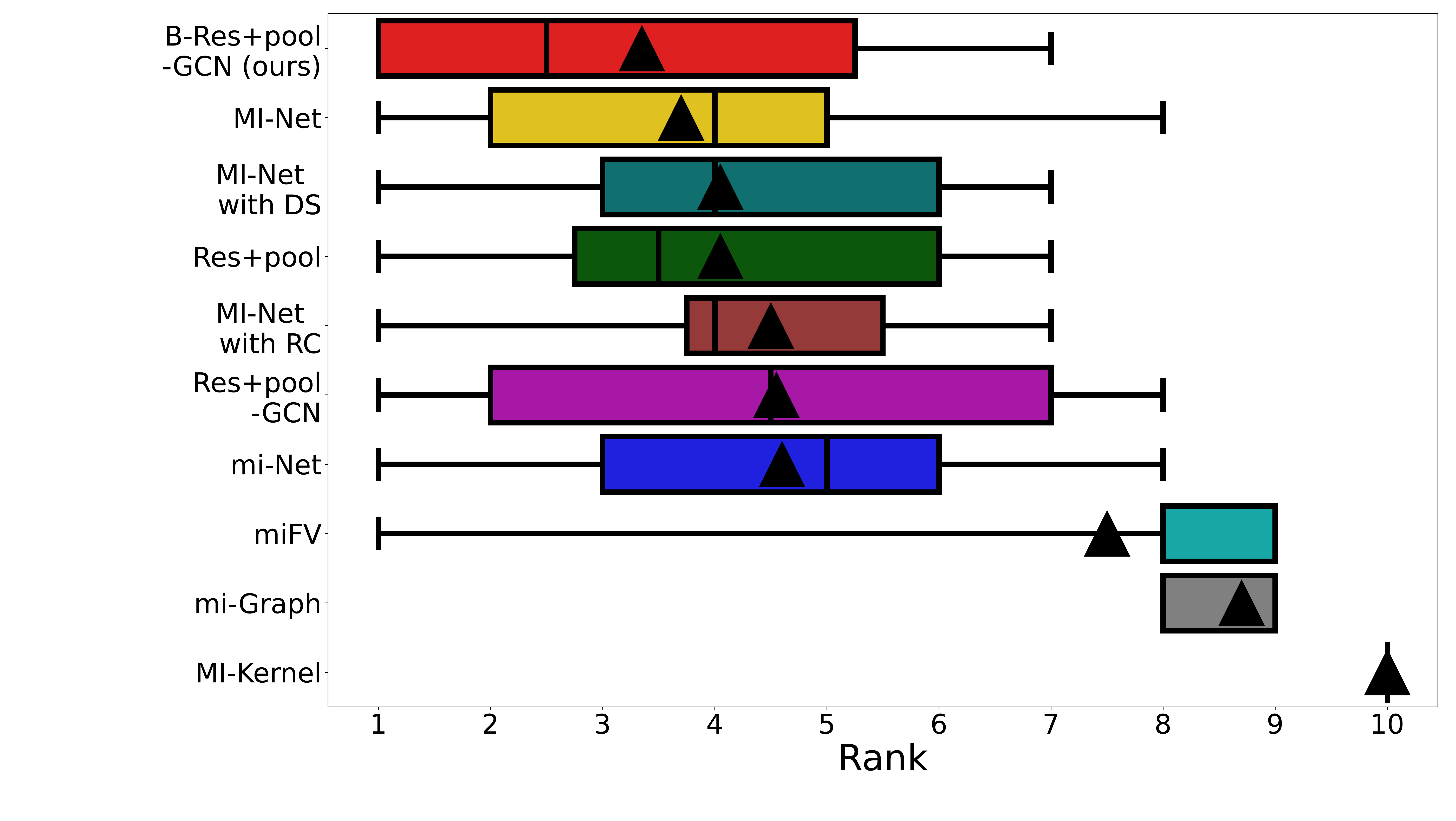}
\vspace{-2.5em}
\caption{Boxplot of ranks of the algorithms across the 20 text datasets. The medians and means of the ranks are shown by the vertical lines and the black triangles respectively; whiskers extend to the minimum and maximum ranks.}
\label{fig:rank_box}
\vspace{-2em}
\end{figure}
\subsection{Text Categorization}
\label{subsec:text}
% \vspace{-0.25em}
We evaluate the proposed approach on 20 text datasets~\cite{zhou2009} derived from the 20 Newsgroup corpus (details in supplementary). 
% Each of these datasets contain 50 positive and 50 negative bags of news articles. Each article (instance) is represented by the top 200 term frequency inverse document frequency (TF-IDF) features. The positive bags contain about 3\% of posts randomly sampled from the target category, whereas the negative bags are made of instances drawn from other categories. The goal is to learn to predict whether a bag contains the posts from the target category or not.  

Aside from the classical MIL models such as MI-Kernel, mi-Graph and mi-FV, we also consider the mi and MI-net models as baselines, as they are shown to outperform the classical models on these datasets in~\citep{wang2018}. 

\begin{figure*}[htbp]
\centering
  \includegraphics[trim={1.25em 1em 1em 1em},clip, width=0.9\textwidth]{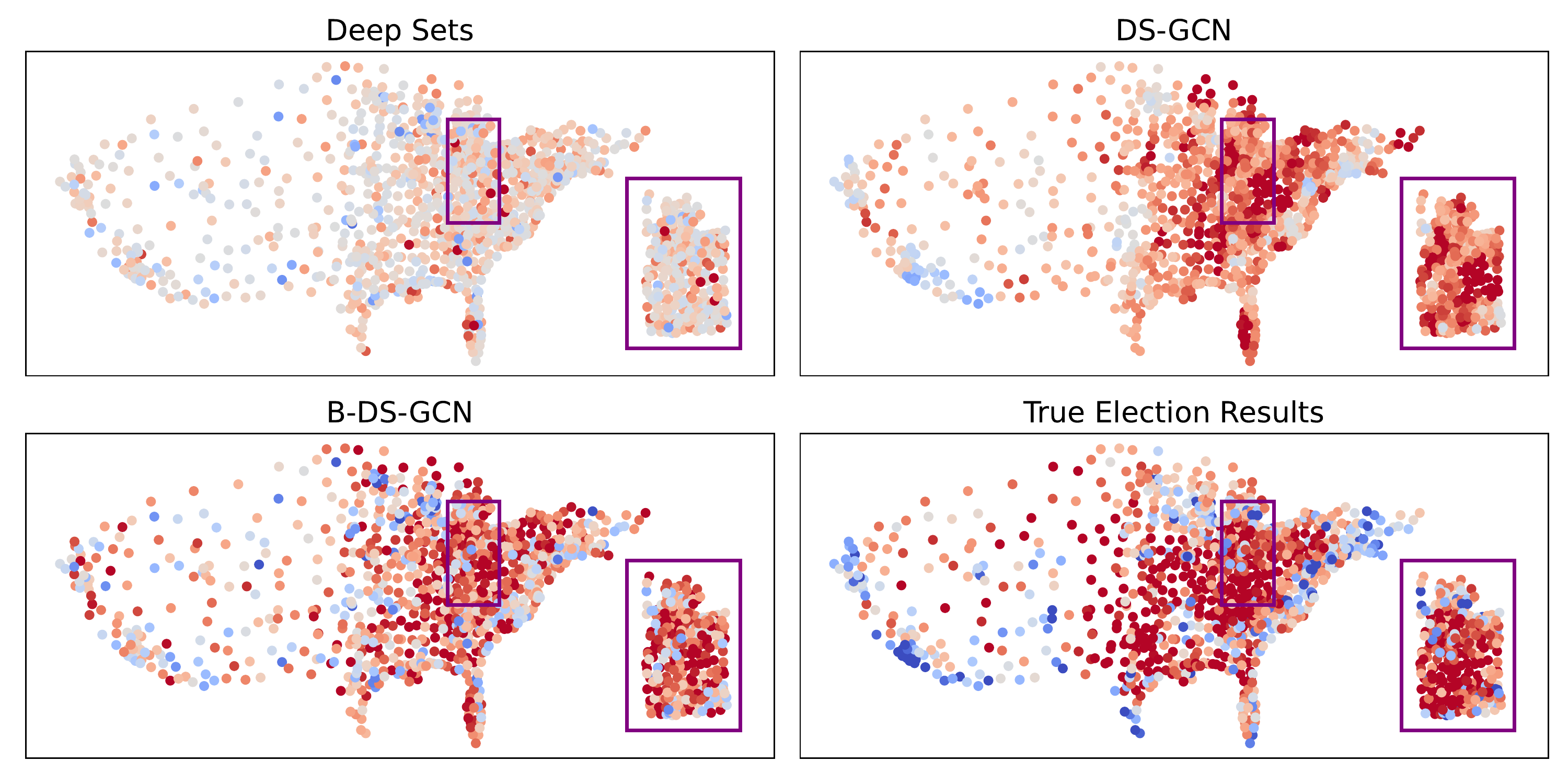}
  \vspace{-0.5em}
  \caption{Predictions of voting probability from Deep Sets, DS-GCN, and B-DS-GCN for the 2016 US presidential election. A county is shown in red (or blue) if the majority votes in favor of republican (or democratic) party. The intensity of the red and blue dots indicates the percentage of the votes obtained by the republican and democratic parties respectively.}
  \label{fig:election}
  \vspace{-2em}
\end{figure*}

% \begin{figure*}[htbp]
%   \includegraphics[width=0.5\textwidth]{figures/rentalListingsNYC.png}
%   \includegraphics[width=0.5\textwidth]{figures/NYCInducedGraph.png}
%   \caption{Real estate dataset visualization. Left: Each red dot represents a rental property in New York City. Right: induced neighborhoods by proximity to official New York City neighborhood centroids. Adapted from: \cite{valkanas2020}.}
%   \label{fig:map}
% \end{figure*}

For this task, we use a residual architecture with pooling as the base model (details in supplementary). We conduct 10 fold cross-validation 10 times using the data-splits of~\citep{zhou2009}. The obtained results are summarized in Table~\ref{tab:text}. The boxplot of the ranks of the algorithms across the 20 datasets is shown in Figure~\ref{fig:rank_box}.
% The detailed architecture and the hyperparameters are listed in the supplementary.

From Table~\ref{tab:text} and Figure~\ref{fig:rank_box}, we observe that all neural network based models outperform the classical MIL models on average in this task. In particular, MI-Net and MI-Net with DS algorithms show impressive performance. We also see that using the $k$-NN heuristic to construct $\mathcal{G}_{obs}$ does not work well for this task, as the Res+pool-GCN algorithm shows worse performance on average compared to the base model. The proposed B-Res+pool-GCN algorithm outperforms the base model considerably and achieves the best average and median ranks among all algorithms across the 20 datasets.
\vspace{-0.75em}
\subsection{Electoral Results Prediction}
\label{subsec:election}
\vspace{-0.25em}
In this task, our aim is to learn to predict the voting pattern of the US counties in the 2016 presidential election. The dataset is obtained from~\citep{flaxman2015} (details in supplementary). In this dataset, people (instances) are associated with the socio-economical features from US census data and we construct the bags by randomly sampling 100 people from each county.

We consider an extreme data-scarce setting, where the data from only 2.5\% of counties (amounting to approximately one county per state) are used for training. We conduct 100 trials where each trial  consists of a random train-test split and random sampling of people to construct the bag feature matrix. $\mathcal{G}_{obs}$ is a $k$-NN graph constructed based on the locations of the centroids of the counties.

We choose Deep Sets (DS)~\cite{deepsets} as a non-graph MIL baseline. Its GCN variant DS-GCN uses $\mathcal{G}_{obs}$ for graph convolution. In order to compute the distance matrix for the non-parametric graph inference step of the proposed Bayesian DS-GCN (B-DS-GCN) algorithm, we use the bag embeddings obtained from the DS-GCN (details in supplementary). We also compare our results with standard MIL baselines, such as MI-Kernel and mi-SVM.
\begin{table}[ht]
\centering
\vspace{-0.75em}
\caption{Average accuracy and ND (in \%) of electoral results prediction reported with std. error over 100 trials.}
 \vspace{-0.5em}
\label{tab:election}
\scriptsize
\setlength{\tabcolsep}{2pt}
\begin{tabular}{lccccc}
\toprule
\textbf{Algorithm} &\textbf{MI-Kernel }&\textbf{mi-SVM } &\textbf{Deep Sets } &\textbf{DS-GCN} &\textbf{B-DS-GCN}\\ \midrule
Accuracy &63.45$\pm$6.10 &72.17$\pm$9.10 &73.22$\pm$3.22 &74.05$\pm$4.56* &\textbf{74.29$\pm$3.15}*  \\
ND  &N/A  &N/A &22.35$\pm$2.66 &21.61$\pm$3.17 &\textbf{21.47$\pm$2.45}  \\
\bottomrule
\end{tabular}
\vspace{-2em}
\end{table}

We conduct a Wilcoxon signed rank test to assess the statistical significance of the obtained results. For the BGCN (or GCN) variant of the base model,  * indicates that the performance of the algorithm is significantly better at the 5\% level compared to the base model. Similarly, ** for the BGCN (or GCN) variant refers to significantly better performance compared to both the base model and its GCN (or BGCN) variant. This procedure is followed in the next experiment as well. 

Table~\ref{tab:election} reports the classification accuracy (republican vs. democrat) and the Normalized Deviation (ND) of the predicted percentage of votes in each county. We observe that the DS-GCN outperforms Deep Sets, since the latter cannot incorporate spatial information. The proposed B-DS-GCN algorithm achieves the highest average accuracy and the lowest average ND. Figure~\ref{fig:election} shows that compared to Deep Sets and DS-GCN, the predicted voting percentages from the proposed B-DS-GCN algorithm show greater agreement to the ground truth. 
\vspace{-0.5em}
\subsection{Rental Price Prediction}
\label{subsec:rental}
\vspace{-0.25em}
As the last task, we evaluate our model in a distribution regression setting to predict mean rental price in New York City neighborhoods. The dataset\footnotemark \footnotetext{\footnotesize{\url{https://www.kaggle.com/c/two-sigma-connect-rental-listing-inquiries/overview}}} includes features of 50,000 rental properties in New York City along with their geographical locations.
Each listing (instance) is described by a vector of features such as a text description, the number of bedrooms, bathrooms, etc. (detailed description and visualization in supplementary).

% A detailed description and visualization of the dataset is provided in the supplementary material

We follow the pre-processing steps described in~\citep{valkanas2020} that include outlier removal, feature standardization, and removing listings with missing attributes. An observed graph ($\mathcal{G}_{obs}$) of 77 nodes is created using data from the official New York City neighborhood map\footnote{\footnotesize{\url{https://data.cityofnewyork.us/City-Government/Neighborhood-Names-GIS/99bc-9p23}}}. Each neighborhood is defined as a node in the graph and we connect nodes with edges based on whether the neighborhoods share a border. The bags are constructed by randomly sampling 25 listings from each neighborhood. For each bag, the label is the true mean of rent prices for all listings within that neighborhood.
% In our experiments the number of listings is set to $n=25$.

We use Root Mean Squared Error (RMSE), Mean Absolute Error (MAE), and Mean Absolute Percentage Error (MAPE) of the predicted average rent as evaluation metrics. We repeat the experiment 100 times using a random 70\%-30\% train-test split of the bags and random sampling of the listings in those bags in each trial. We consider Deep Sets (DS)~\cite{deepsets} and Set Transformer (ST)~\cite{set_transformer} as the two graph agnostic baselines for this regression task. Their GCN variants DS-GCN and ST-GCN and BGCN variants B-DS-GCN and B-ST-GCN are constructed as in the previous experiment (details in supplementary).

% {\color{red} We consider Deep Sets (DS)~\cite{deepsets} and Set Transformer (ST)~\cite{set_transformer} as two graph agnostic baselines for this regression task. Their GCN variants DS-GCN and ST-GCN use $\mathcal{G}_{obs}$ for graph convolution. In order to compute the distance matrix for the non-parametric graph inference step of the proposed Bayesian DS-GCN (B-DS-GCN) and Bayesian ST-GCN (B-ST-GCN) algorithms, we use the bag embeddings obtained from the corresponding GCN variants. The detailed architectures of these models and other hyperparameters are listed in the supplementary.}

The results are summarized in Table~\ref{tab:rental}. We observe that both DS-GCN and ST-GCN outperform their corresponding base models significantly, which shows that the utilization of spatial information encoded by $\mathcal{G}_{obs}$ is beneficial for the task. The proposed B-DS-GCN and B-ST-GCN provide further improvement in almost all cases. This suggests that beyond simple geographical proximity, the proposed approach is capable of learning more complex relationships among the neighborhoods influencing the mean rental prices. 
\begin{table}[ht]
\centering
\vspace{-0.75em}
\caption{Average RMSE, MAE, and MAPE for rental price prediction reported with std. error over 100 trials. %The best and the second best results in each column are shown in bold and marked with underline respectively.
}
\label{tab:rental}
\vspace{-0.55em}
\scriptsize
\setlength{\tabcolsep}{2pt}
\begin{tabular}{lccc}
\toprule
\textbf{Algorithm}  &\textbf{RMSE}  &\textbf{MAE} &\textbf{MAPE (\%)}\\ \midrule
Deep Sets &86.37$\pm$20.41 &65.19$\pm$15.72 &2.24$\pm$0.36 \\
DS-GCN  &78.57$\pm$16.06* &59.21$\pm$10.20* &1.92$\pm$0.24* \\
B-DS-GCN &\textbf{67.51$\pm$16.39}** &\textbf{47.24$\pm$10.21}** &\underline{1.83$\pm$0.20}** \\ \midrule
Set Transformer &76.34$\pm$15.04 &56.09$\pm$9.10 &2.02$\pm$0.22 \\
ST-GCN &71.86$\pm$14.65* &53.56$\pm$9.11* &\textbf{1.81$\pm$0.22}*\\
B-ST-GCN &\underline{69.44$\pm$16.23}** &\underline{49.72$\pm$9.60}** &\underline{1.83$\pm$0.22}* \\ 
\bottomrule
\end{tabular}
\vspace{-2.5em}
\end{table}

\begin{table}[ht]
\centering
\vspace{-0.25em}
\caption{Ablation study for rental price prediction: average RMSE, MAE, and MAPE with std. error over 100 trials.}
\label{tab:rental_abltion}
\vspace{-0.65em}
\scriptsize
\setlength{\tabcolsep}{1pt}
\begin{tabular}{llccc}
\toprule
&\textbf{Algorithm}  &\textbf{RMSE}  &\textbf{MAE} &\textbf{MAPE (\%)}\\ \midrule
{\multirow{3}{*}{\rotatebox[origin=c]{90}{  DS }}}&\text{ }t. n. d. &75.26$\pm$16.99 &54.48$\pm$12.01 &2.03$\pm$0.25 \\
&\text{ }t. n. d. during training   &68.15$\pm$16.77* &48.08$\pm$10.77* &1.85$\pm$0.22* \\
&\text{ }transductive &67.51$\pm$16.39* &47.24$\pm$10.21* &1.83$\pm$0.20** \\ \midrule
{\multirow{3}{*}{\rotatebox[origin=c]{90}{  ST }}}&\text{ }t. n. d. &89.95$\pm$23.23 &67.12$\pm$18.34 &2.29$\pm$0.47 \\
&\text{ }t. n. d. during training &71.72$\pm$16.50* &51.66$\pm$10.23* &1.88$\pm$0.26* \\
&\text{ }transductive &69.44$\pm$16.23** &49.72$\pm$9.60** &1.83$\pm$0.22** \\
\bottomrule
\end{tabular}
\vspace{-1.5em}
\end{table}
We conduct an ablation study to determine if the transductive setting employed in this work is indeed beneficial. For both architectures, we consider a scenario with `test nodes disconnected' (t. n. d.), which refers to the case where graph inference is carried out for the training nodes only, and  disconnected test nodes are added to the graph of training nodes during testing. The other setting is `test nodes disconnected during training', where the training is carried out based on the inferred graph of training nodes, but the learned model is evaluated on the inferred graph of both training and test set nodes. From the results in Table~\ref{tab:rental_abltion}, we note that both conducting the non-parametric graph inference for training and test set nodes together and training the model in a transductive setting contribute positively to the outcome of this task.
\vspace{-2.25em}
\section{Conclusion}
\vspace{-0.5em}
In this paper, we have proposed a novel graph-based MIL method that is capable of addressing learning problems where there is relational information between the bags to be labeled. We employ a Bayesian graph neural network framework which allows a graph to be inferred from the data, so our method is also applicable to the traditional MIL setting where no graph is specified. The proposed methodology is generally applicable to diverse MIL problem settings, as it can incorporate various existing deep learning based MIL models to learn bag representations and aggregate them using a Bayesian GNN via end-to-end training. Empirical results demonstrate that the proposed method achieves performance comparable to the state-of-the-art on MIL benchmark datasets, and offers better performance in text categorization, electoral results prediction, and rental price regression. Some potential future research directions include adapting the methodology to the inductive setting by using inductive GNN variants~\cite{hamilton2017} and improving the training efficiency of the overall architecture by using node or graph sampling~\cite{chiang2019, zeng2020}.

\section*{Acknowledgements}
We acknowledge the support of the Natural Sciences and Engineering Research Council (NSERC) of Canada, [funding reference number 260250].

\bibliography{refs}
\end{document}

\maketitle

\section{Experimental Details} 
\label{sec:experiment_details}
We provide details of the datasets and experimental setups for all tasks considered in the main paper.
\subsection{Classification of benchmark MIL datasets}
 \textbf{Datasets:} The statistics of the five MIL benchmark datasets are provided in Table~\ref{tab:mil_description}. The MUSK1 and MUSK2~\cite{musk} datasets are composed of 92 and 102 molecules, respectively. Some of the molecules have a musky smell while others do not. The goal is to predict whether the unlabelled molecules are musky or non-musky. A bag is constructed for each molecule with its various geometric arrangements, called conformations, as instances. Each instance is represented with a 166 dimensional feature vector. If a molecule has at least one musky conformation, it is labeled as musky. The FOX, TIGER, and ELEPHANT~\cite{andrews2002} datasets consist of 200 bags of features extracted from animal images. A bag is labeled positive if at least one instance contains the relevant animal whereas the negative bags contain images of different animals. In these datasets, each image is associated with a 230 dimensional feature vector. 

\begin{table}[ht]
\caption{Statistics of the MIL benchmark datasets.}
\label{tab:mil_description}
\scriptsize
\setlength{\tabcolsep}{4pt}
\centering
\begin{tabular}{lccccc}
\hline
\multicolumn{1}{c}{\textbf{Dataset}} & \textbf{MUSK1} & \textbf{MUSK2} & \textbf{FOX} & \textbf{TIGER} & \textbf{ELEPHANT} \\ \hline
No. features                           & 166            & 166            & 230          & 230            & 230               \\ 

No. total bags                         & 92             & 102            & 200          & 200            & 200               \\ 
No. positive bags                      & 47             & 39             & 100          & 100            & 100               \\ 
No. negative bags                      & 45             & 63             & 100          & 100            & 100               \\ 
Min. instances in a bag                & 2              & 1              & 2            & 1              & 3                 \\
Max. instances in a bag                & 40             & 1044           & 13           & 13             & 13                \\ 
No. total instances                    & 476            & 6598           & 1320         & 1220           & 1391              \\ \hline
\end{tabular}
\end{table}

\textbf{Architecture and hyperparameters:} We use a 3-layer row-wise FeedForward (rFF) architecture with 256, 128, and 64 hidden units respectively. Each layer has a ReLU activation function. Inputs to the deep supervision layers~\cite{wang2018} are subjected to dropout with probability 0.5. For the rFF+pool-GCN and B-rFF+pool-GCN models, these linear deep supervision layers are replaced by GCN~\cite{kipf2017} layers. Using a 10-fold cross validation, we select the number of neighbours $k$ for the $k$-NN graphs  by searching over $k \in \{1, 2, 3, 4\}$.
This search procedure is used for $\mathcal{G}_{obs}$ in rFF+pool-GCN and for the estimated graph $\widehat{\mathcal{G}}$ in B-rFF+pool-GCN. The other hyperparameter, $r$, that is associated with obtaining $\widehat{\mathcal{G}}$, is chosen from $\{1, 5, 10\}$. The identified values for various hyperparameters for each dataset are summarized in Table~\ref{tab:mil_hyperparam}. We observe that the cross-validation selects $k=1$ for both rFF+pool-GCN and B-rFF+pool-GCN algorithms on the ELEPHANT dataset. This shows that use of graphs is not particularly useful for this dataset. We use the Adam optimizer to minimize training set cross-entropy loss for 200 epochs.

\begin{table}[ht]
\caption{Hyperparameters for the MIL benchmark datasets.}
\label{tab:mil_hyperparam}
\scriptsize
\setlength{\tabcolsep}{4pt}
\centering
\begin{tabular}{lccccc}
\hline
\multicolumn{1}{c}{\textbf{Dataset}} & \textbf{MUSK1} & \textbf{MUSK2} & \textbf{FOX} & \textbf{TIGER} & \textbf{ELEPHANT} \\ \hline
Learning rate                           &0.0005            &0.0005            &0.0001         &0.0005            &0.0001               \\ 

Weight decay                         &0.005             &0.03           &0.01          &0.005            &0.005               \\
Pooling method                         &\emph{max}             &\emph{max}            &\emph{max}          &\emph{mean}            &\emph{max}               \\
$k$ in rFF+pool-GCN                      &2            &3             &3          &4            &1               \\ 
$k$ in B-rFF+pool-GCN                      &2             &3             &3          &4            &1               \\ 
$r$ in B-rFF+pool-GCN                &1              &10              &5           &10             &10                 \\ \hline
\end{tabular}
\end{table}

% \MJC{Maybe add some text discussing the identified values and the relationship with performance? (if there is one?) As you said, Elephant =1 seems to suggest graph is not very useful
% } 

\subsection{Text Categorization}
\textbf{Datasets:} The detailed statistics of the 20 text datasets~\cite{zhou2009} derived from the 20 Newsgroup corpus are summarized in Table~\ref{tab:text_description}. 
Each of these datasets contain 50 positive and 50 negative bags of news articles. Each article (instance) is represented by the top 200 term frequency inverse document frequency (TF-IDF) features. The positive bags contain about 3\% of posts randomly sampled from the target category, whereas the negative bags are made of instances drawn from other categories. The goal is to learn to predict whether a bag contains the posts from the target category or not.  

\begin{table}[ht]
\caption{Statistics of the 20Newsgroup datasets.}
\label{tab:text_description}
\scriptsize
\setlength{\tabcolsep}{4pt}
\centering
\begin{tabular}{lccc}
\hline
\textbf{Dataset}         & \textbf{\begin{tabular}[c]{@{}c@{}}Min. instances\\ in a bag\end{tabular}} & \textbf{\begin{tabular}[c]{@{}c@{}}Max. instances\\ in a bag\end{tabular}} & \textbf{\begin{tabular}[c]{@{}c@{}}No. total\\ instances\end{tabular}} \\ \hline
\emph{alt.atheism}              & 22                                                                         & 76                                                                         & 5443                                                                   \\ 
\emph{comp.graphics}            & 12                                                                         & 58                                                                         & 3094                                                                   \\ 
\emph{comp.os.ms-windows.misc}  & 25                                                                         & 82                                                                         & 5175                                                                   \\ 
\emph{comp.sys.ibm.pc.hardware} & 19                                                                         & 74                                                                         & 4287                                                                   \\ 
\emph{comp.sys.mac.hardware}    & 17                                                                         & 71                                                                         & 4473                                                                   \\ 
\emph{comp.windows.x}           & 12                                                                         & 54                                                                         & 3110                                                                   \\ 
\emph{misc.forsale}             & 29                                                                         & 84                                                                         & 5306                                                                   \\ 
\emph{rec.autos}                & 15                                                                         & 59                                                                         & 3458                                                                   \\ 
\emph{rec.motorcycles}          & 22                                                                         & 73                                                                         & 4730                                                                   \\ 
\emph{rec.sport.baseball}       & 15                                                                         & 58                                                                         & 3358                                                                   \\ 
\emph{rec.sport.hockey}         & 8                                                                          & 38                                                                         & 1982                                                                   \\ 
\emph{sci.crypt}                & 20                                                                         & 71                                                                         & 4284                                                                   \\ 
\emph{sci.electronics}          & 12                                                                         & 58                                                                         & 3192                                                                   \\ 
\emph{sci.med}                  & 11                                                                         & 54                                                                         & 3045                                                                   \\ 
\emph{sci.space}                & 16                                                                         & 59                                                                         & 3655                                                                   \\ 
\emph{soc.religion.christian}   & 21                                                                         & 71                                                                         & 4677                                                                   \\ 
\emph{talk.politics.guns}       & 13                                                                         & 59                                                                         & 3558                                                                   \\ 
\emph{talk.politics.mideast}    & 15                                                                         & 55                                                                         & 3376                                                                   \\ 
\emph{talk.politics.misc}       & 21                                                                         & 75                                                                         & 4778                                                                   \\ 
\emph{talk.religion.misc}       & 25                                                                         & 79                                                                         & 4606                                                                   \\ \hline
\end{tabular}
\end{table}

\textbf{Architecture and hyperparameters:} We use a 3-layer residual architecture with 128 hidden units and ReLU activation function for the Res+pool model. Dropout with probability 0.5 is applied to the last representation layer. For the Res+pool-GCN and B-Res+pool-GCN models, we replace this layer by a GCN~\cite{kipf2017} layer. Mean pooling is chosen for all datasets as it consistently outperforms the max pooling in this experiment. For both Res+pool-GCN and B-Res+pool-GCN algorithms, $k$ is chosen from $\{2,3,4\}$. The other hyperparameter of B-Res+pool-GCN, $r$, is selected from $\{1, 5, 10\}$. The chosen values for these hyperparameters are listed in Table~\ref{tab:text_hyperparam}. All models are trained for 200 epochs to minimize binary cross-entropy on the training set using the Adam optimizer with learning rate 0.001 and weight decay 0.001. 

\begin{table}[ht]
\caption{Graph-related hyperparameters for the 20Newsgroups datasets}
\label{tab:text_hyperparam}
\scriptsize
\setlength{\tabcolsep}{4pt}
\centering
\begin{tabular}{lccc}
\hline
\textbf{Dataset}         & \textbf{\begin{tabular}[c]{@{}c@{}}$k$ in Res+\\ pool-GCN\end{tabular}} & \textbf{\begin{tabular}[c]{@{}c@{}}$k$ in B-Res+ \\pool-GCN\end{tabular}} & \textbf{\begin{tabular}[c]{@{}c@{}}$r$ in B-Res+ \\pool-GCN\end{tabular}} \\ \hline
\emph{alt.atheism}              &2                                                                         &3                                                                         &10                                                                  \\ 
\emph{comp.graphics}            &2                                                                         &3                                                                         &5                                                                   \\ 
\emph{comp.os.ms-windows.misc}  &2                                                                         &3                                                                         &10                                                                   \\ 
\emph{comp.sys.ibm.pc.hardware} &3                                                                         &4                                                                         &5                                                                   \\ 
\emph{comp.sys.mac.hardware}    &3                                                                         &3                                                                         &5                                                                   \\ 
\emph{comp.windows.x}           &4                                                                         &3                                                                         &10                                                                   \\ 
\emph{misc.forsale}             &3                                                                         &3                                                                         &1                                                                   \\ 
\emph{rec.autos}                &4                                                                         &2                                                                         &1                                                                   \\ 
\emph{rec.motorcycles}          &2                                                                         &3                                                                         &10                                                                   \\ 
\emph{rec.sport.baseball}       &2                                                                         &2                                                                         &10                                                                   \\ 
\emph{rec.sport.hockey}         &3                                                                          &4                                                                         &1                                                                   \\ 
\emph{sci.crypt}                &3                                                                         &3                                                                         &10                                                                   \\ 
\emph{sci.electronics}          &3                                                                         &2                                                                         &1                                                                   \\ 
\emph{sci.med}                  &3                                                                         &3                                                                         &10                                                                   \\ 
\emph{sci.space}                &4                                                                         &4                                                                         &5                                                                   \\ 
\emph{soc.religion.christian}   &3                                                                         &4                                                                         &10                                                                   \\ 
\emph{talk.politics.guns}       &2                                                                         &3                                                                         &10                                                                   \\ 
\emph{talk.politics.mideast}    &2                                                                         &2                                                                         &5                                                                   \\ 
\emph{talk.politics.misc}       &4                                                                         &3                                                                         &10                                                                   \\ 
\emph{talk.religion.misc}       &2                                                                         &4                                                                         &5                                                                   \\ \hline
\end{tabular}
\end{table}

\begin{figure*}[htbp]
  \includegraphics[width=0.5\textwidth]{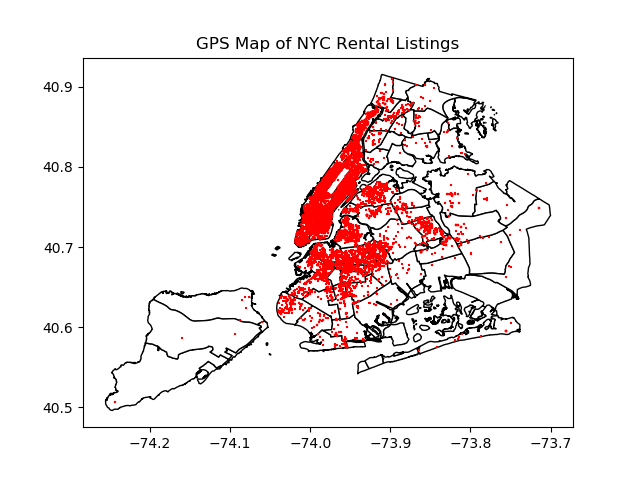}
  \includegraphics[width=0.5\textwidth]{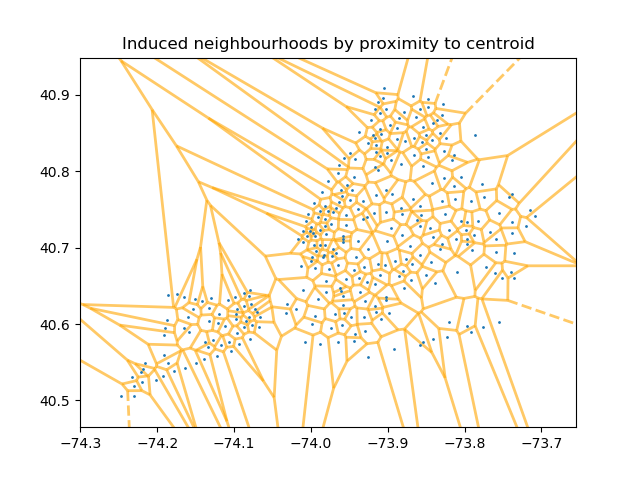}
  \caption{Real estate dataset visualization. Left: Each red dot represents a rental property in New York City. Right: induced neighborhoods by proximity to official New York City neighborhood centroids. Adapted from: \cite{valkanas2020}.}
  \label{fig:map}
\end{figure*}

\subsection{Electoral Results Prediction}
\textbf{Dataset:} The dataset is obtained from~\citep{flaxman2015}. In this dataset\footnotemark \footnotetext{\footnotesize{\url{https://github.com/flaxter/us2016}}}, there are 979 counties and people (instances) are associated with the socio-economical features from US census data. After removing the attributes with missing values, each instance is associated with a 94 dimensional feature vector. We construct the bags by randomly sampling 100 people from each county. The bag labels are the voting percentage for the Republican and Democratic parties in each county.

\textbf{Architecture and hyperparameters:} We use Deep Sets~\cite{deepsets} as the base model in this task. The instances are fed to a 2-layer FeedForward architecture with 128 hidden units and ReLU activation. The resulting instance representations are summed within each set and we use a 2-layer bag representation learning module with hidden dimensions 128 and 64 and ReLU activation for forming the final prediction. The last linear layer is replaced by a GCN~\cite{kipf2017} layer for the DS-GCN and the B-DS-GCN algorithms. For constructing the $k$-NN graph $\mathcal{G}_{obs}$ from the location of the county centroids, we use $k=5$. For the proposed B-DS-GCN, we set $k=5$ and $r=1$. All models are trained for 200 epochs using the Adam optimizer to minimize training set cross-entropy loss. The learning rate is 0.001 and the weight decay is set to 0.0001.

\subsection{Rental Price Prediction}
\textbf{Dataset:} The NYC rental dataset\footnotemark \footnotetext{\footnotesize{\url{https://www.kaggle.com/c/two-sigma-connect-rental-listing-inquiries/overview}}} includes features of 50,000 rental properties in New York City along with their geographical locations.
Each listing (instance) is described by a vector of features such as a text description, the number of bedrooms, bathrooms, etc. Figure~\ref{fig:map} provides a  visualization of the locations of the properties.

We follow the pre-processing steps described in~\citep{valkanas2020} that include outlier removal, feature standardization, and removing listings with missing attributes. Each instance is characterized by a 10 dimensional feature vector. An observed graph ($\mathcal{G}_{obs}$) of 77 nodes is created using data from the official New York City neighborhood map\footnotemark\footnotetext{\footnotesize{\url{https://data.cityofnewyork.us/City-Government/Neighborhood-Names-GIS/99bc-9p23}}}. Each neighborhood is defined as a node in the graph and we connect nodes with edges based on whether the neighborhoods share a border. The bags are constructed by randomly sampling 25 listings from each neighborhood. For each bag, the label is the true mean of rent prices for all listings within that neighborhood.

\textbf{Architecture and hyperparameters:} For the Deep Sets~\cite{deepsets} model, we use a 3-layer architecture with 25 hidden units and ELU activation for instance representation learning. The bag representation learning module takes the sum of the instance representation within a set as input and applies another 4-layer feed-forward architecture with 25 hidden units and ELU activation at each hidden layer to obtain 64 dimensional bag embeddings. The Set Transformer~\cite{set_transformer} architecture consists of a 64 dimensional Set Attention Block (SAB) layer, followed by a PMA (Pooling by Multihead Attention)  layer of the same dimension. In both cases, the GCN and BGCN variants are constructed by replacing the last linear layer by a GCN layer. For the Bayesian approaches, we set $k=8$, which is also the average degree of the nodes in the observed graph $\mathcal{G}_{obs}$. We set the other hyperparameter $r=1$. For this task, we use the MSE of the predictions on the training set as the loss function and minimize it using the Adam optimizer for 500 epochs. The learning rate is set to 0.0005 and the weight decay is 0.001.

\bibliography{refs}